# Advancing breast, lung and prostate cancer research with federated learning. A systematic review

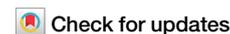
Check for updates

Anshu Ankolekar[1] ✉, Sebastian Boie[2], Maryam Abdollahyan[3], Emanuela Gadaleta[3], Seyed Alireza Hasheminasab[4], Guang Yang[5], Charles Beauville[6], Nikolaos Dikaios[7], George Anthony Kastis[7], Michael Bussmann[8], Claude Chelala[3], Sara Khalid[4], Hagen Kruger[2], Philippe Lambin[1] & Giorgos Papanastasiou[9] ✉ on behalf of OPTIMA Consortium*

Federated learning (FL) is advancing cancer research by enabling privacy-preserving collaborative training of machine learning (ML) models on diverse, multi-centre data. This systematic review synthesises current knowledge on state-of-the-art FL in oncology, focusing on breast, lung, and prostate cancer. Unlike previous surveys, we critically evaluate FL's real-world implementation and impact, demonstrating its effectiveness in enhancing ML generalisability and performance in clinical settings. Our analysis reveals that FL outperformed centralised ML in 15 out of 25 studies, spanning diverse models and clinical applications, including multi-modal integration for precision medicine. Despite challenges identified in reproducibility and standardisation, FL demonstrates substantial potential for advancing cancer research. We propose future research focus on addressing these limitations and investigating advanced FL methods to fully harness data diversity and realise the transformative power of cutting-edge FL in cancer care.

Oncology is undergoing rapid transformation due to the integration of machine learning (ML), which can enrich clinical evidence from large-scale datasets, surpassing traditional analytics[1–4]. However, as of today, ML models have predominantly been centralised within data silos[1,2]. While centralised ML models have substantially advanced cancer research[3], the exponential growth and diversification of clinical data such as imaging, health records and molecular profiles now pose considerable challenges[4]. This surge in data, coupled with a trend toward international collaboration and standardised datasets, highlights the limitations of single-centre studies confined by local data acquisition practices and demographics. Multi-centre studies, drawing from diverse regions, offer a more comprehensive ML modelling approach. However, centralised models struggle to effectively exploit this increasingly complex data landscape, potentially compromising ML generalisability, performance, global applicability and trustworthiness[5]. While aggregating data from various sources in centralised data lakes potentially offers an alternative, it is susceptible to privacy breaches, complex data-sharing agreements and legal restrictions on data transfers[1].

Federated learning (FL) has emerged as a potential solution to these limitations. With FL, ML algorithms can be trained simultaneously on local datasets without any direct data exchange between participating entities[6]. In FL, a central server distributes an initial model to participating institutions. Each institution trains this model locally on its own data and transmits only the model updates (e.g., weights, gradients) back to the server. The server aggregates these updates to create an improved global model, which is then redistributed for further rounds of training. The iterative process continues until the global model reaches a desired level of performance. This decentralised approach allows hospitals and institutes to retain control over their data, addressing privacy concerns and regulatory restrictions, while benefiting from collective insights[7]. Therefore, FL enables access to the large, diverse datasets crucial for advancing clinical decisions in precision medicine. For instance, federated random forests have offered a privacy-preserving solution for leveraging distributed data to improve the accuracy of local models[8].

[1]Department of Precision Medicine, GROW Research Institute for Oncology and Reproduction, Maastricht University, Maastricht, the Netherlands. [2]Pfizer Pharma GmbH, Berlin, Germany. [3]Centre for Cancer Biomarkers and Biotherapeutics, Barts Cancer Institute, Queen Mary University of London, London, United Kingdom. [4]Centre for Statistics in Medicine, Nuffield Department of Orthopaedics, Rheumatology and Musculoskeletal Sciences, University of Oxford, Oxford, United Kingdom. [5]Bioengineering Department and Imperial-X, Imperial College London, London, UK. [6]Flower Labs, Hamburg, Germany. [7]Mathematics Research Center, Academy of Athens, Athens, Greece. [8]Helmholtz-Zentrum Dresden-Rossendorf, Dresden, Germany. [9]Artificial Intelligence Data and Analytics, Digital, Pfizer Inc, New York, USA. *A list of authors and their affiliations appears at the end of the paper. ✉e-mail: a.ankolekar@maastrichtuniversity.nl; georgios.papanastasiou@pfizer.com

FL is particularly promising in oncology, where the data pertains to sensitive patient information and where timely collaborative analysis can have a significant impact on diagnostic performance and patient outcomes[9]. Recent reviews highlight the growing importance of FL in oncology by exploring its applications, technical architectures, and future directions, providing a valuable foundation for understanding its potential in healthcare[10,11]. However, there remains a critical need for a thorough evaluation of FL's real-world implementation and impact within specific oncological contexts, including an assessment of its effectiveness in enhancing model generalizability and performance. In addition, the adoption of FL is not without its challenges. Balancing effective model training with patient privacy techniques that can add computational overheads and may affect data contents, ensuring data quality and consistency across multiple centres, and maintaining robust model performance and trustworthiness are pressing concerns[12].

Given the rapid evolution and potential of FL in oncology, we conducted a systematic review to synthesise current knowledge, identify best practices, and highlight gaps in the existing literature. This review will provide researchers and clinicians with a comprehensive understanding of how state-of-the-art FL can be effectively implemented to overcome the limitations of centralised ML. We analyse key FL/ML architectural and implementation designs, critically evaluate the effectiveness and scope of FL in breast, lung, and prostate cancer, and discuss best practices and considerations for future work. Further, we assess FL rigour using 2 objective criteria: a) inclusion of a comparative framework to evaluate the proposed FL/ML model against centralised ML baselines developed on the same datasets (either through direct evaluation in the study or by referencing reported literature values), and/or b) whether the proposed FL/ML model surpasses or demonstrates comparable performance to these baselines.

We evaluated state-of-the-art advances in FL, demonstrating its growing adoption amid tightening data privacy regulations. We perform this review analysis in the framework of the OPTIMA IMI2 project (OPTIMA|IMI Innovative Medicines Initiative), which focuses on combining ML and FL to enhance personalised diagnosis and treatment in breast, lung, and prostate cancer, and forms a systematic initiative to address global clinical challenges and unmet needs in cancer research. We aim to provide a comprehensive review of the current state of cutting-edge FL in these 3 oncology areas, to inform future research and clinical practices in the framework of the OPTIMA consortium and beyond.

## Results
### Research trends
Following the Preferred Reporting Items for Systematic Reviews and Meta-Analyses (PRISMA) guidelines[13], our review of FL publications in breast, lung, and prostate cancer research from 2020 to 2023 reveals a growing interest, with none, 4, 8 and 13 papers identified respectively[14–38].

We observed a diverse exploration of ML techniques including Large Pre-trained (10 papers), UNet (7), CNN (6), Classic ML (6), GAN (2), and Other (3). The majority addressed classification (14 papers), followed by segmentation (5), detection (5), and regression (1) (Fig. 1b). Table 1 details technical tasks, ML algorithms, and evaluation metrics across studies.

In terms of clinical application, most studies focused on tumour identification (8 papers) and disease type differentiation (7), followed by severity assessment (4), treatment response prediction (2), survival analysis (2), and Other (2) (Fig. 2a). Five studies did not mention their clinical application. On FL scope, the majority of FL techniques focused on improving ML model generalisability (14 papers), followed by ML prediction improvement (13), data privacy (8), disease understanding improvement (2), and Other (2) (Fig. 2b). Table 2 details FL scopes and oncology areas.

Supplementary Table 1 details the 25 studies included in our systematic review, summarizing key data characteristics such as data source (private/public), patient population size, and data size.

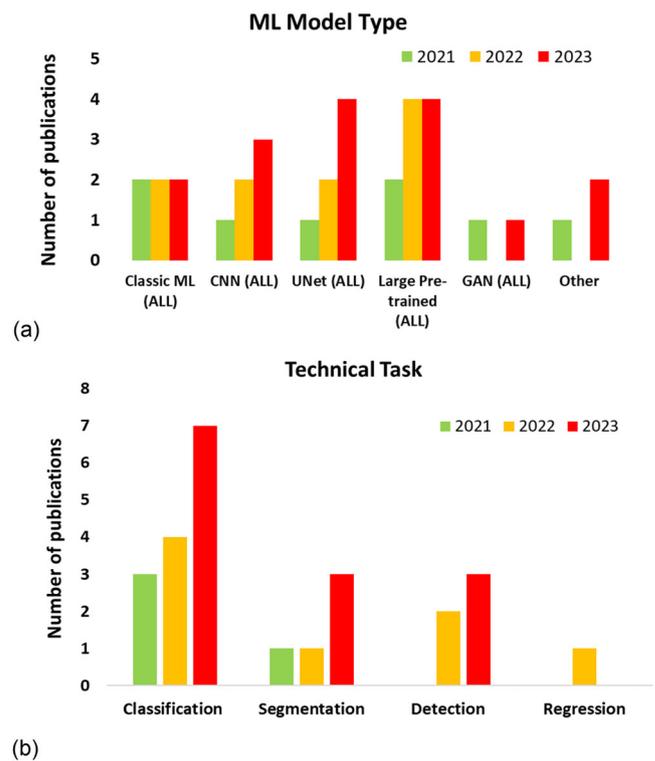

Fig. 1 | Vertical bar graphs showing research trends for all categories of machine learning (ML) models and technical tasks, identified. Publication record over time for all ML types (**a**) and technical tasks (**b**) identified. In **a**, "Other" included recurrent neural network, capsule neural network and region-based CNN.

### Federated machine learning
We analysed combinations of datasets, central ML models, technical tasks, and FL methods (Fig. 3). The Large Pre-trained models were used for (bio-)medical imaging data: magnetic resonance imaging (MRI), computed tomography (CT), whole-slide imaging (WSI), and X-ray. The UNet models were developed for MRI, CT, hybrid positron emission tomography-CT (PET-CT), and X-ray. The CNN models analysed diverse datasets including WSI, digital mammography, MRI, CT, and electronic health records (EHR). The Classic ML models were primarily used for EHR and imaging features extracted from CT[14,25] and MRI[25]. GAN models were only developed for MRI and CT[22,25] and Other models for EHR, CT, and digital mammography[21,34].

All ML model types addressed classification, with Large Pre-trained, Classic ML, and CNN (ALL), being the 3 most frequently used models (Fig. 3, Table 1). Classification tasks were mainly addressed for (bio-)medical imaging data followed by EHR (Fig. 3)[19,27,29,30,32,33,35–38]. Segmentation and detection tasks were mainly approached through UNet variants[18,25,28,31,37] and Large Pre-trained models[20,23], respectively. Some studies addressed multiple tasks (Table 1).

### Federated cancer research
Subsequently, we analysed combinations of clinical applications, datasets, FL scope, and organ areas (Fig. 4).

Tumour identification was addressed using (bio-)medical imaging data: CT[20,21,25], MRI[25,37], X-ray[28,38], PET-CT[18] and WSI[19]. Disease type differentiation was addressed using X-ray[38], WSI[19,33], digital mammography[32,34] and EHR data[17]. Severity assessment (staging) was investigated through (bio-)medical imaging data[37]. Two pairs of studies explored the combination of tumour identification with other clinical applications: 1 pair focused on disease type differentiation (using WSI and X-ray respectively)[19,38], while the other examined severity assessment (CT and MRI respectively)[21,37]. Treatment response prediction (CT, MRI, WSI)[15] and survival analysis (clinical

Table 1 | ML algorithm and evaluation metrics per technical task are presented

| Technical task | ML algorithm | Evaluation | Studies |
|---|---|---|---|
| Classification | Classic ML | ROC | 14,16,27 |
| | | Accuracy | 17,29 |
| | | Other (MCC, Precision, Recall, F1) | 17,30 |
| | CNN | ROC | 15,32,36 |
| | | Other (Precision-Recall AUC, F1, Accuracy) | 32,36 |
| | Large Pre-trained | ROC | 19,23,35 |
| | | F1 score | 24,38 |
| | | Accuracy | 23,24,33,35,38 |
| | | Other (Confusion matrices, Precision, Recall, Kappa coefficient) | 24,33,35,38 |
| | GAN | ROC | 22,24 |
| Detection | Large Pre-trained | ROC | 20,23,34 |
| | Other | Other (Precision, F1, MCC) | 34 |
| Segmentation | UNet | Dice coefficient | 18,25,31,37 |
| | | Other (IoU) | 18,37 |
| Regression | Other (CNN) | Other (Dice coefficient, Accuracy) | 26,28 |

Publications occurring in multiple cells represent overlapping tasks and/or evaluation metrics. In papers where multiple (> 1) tasks were involved, we only present the ML model for which evaluation metrics are reported. Note that we refer to technical tasks, ML algorithms and evaluation metrics, as reported in each publication. The term "Accuracy" presents a standalone metric (separate to ROC analysis), as reported by each paper authors. The term "Other" represents metrics that occurred once per ML model/technical task. ML: machine learning; ROC: receiver operating characteristic curve; MCC: Matthew's correlation coefficient; AUC: area under the curve; IoU: intersection-over-union.

data, genomics)[30] were also examined as standalone tasks, and in combination (EHR, CT)[16]. Side effect prediction (EHR, CT)[14] and tumour recurrence (EHR)[29] were examined in 1 paper each.

In terms of FL scopes, most articles focused on model generalisability and ML prediction improvement (Figs. 2b and 4, Table 2). For both ML model generalisability and prediction improvement, tumour identification was the predominant task (Fig. 4). Breast and lung cancer research were the most frequent (8 studies each, Table 2).

### Data diversity
The size of patient cohorts and data volumes varied considerably (Table 3). Most studies had cohorts of 100–1500 patients, with data sizes (images or samples) in the range 1–5000[15,21] and 5000–1,000,000[20,30,31]. Six papers analysed data from ≥1500 patients, with 3 of these having explored data from >10,000[16,27] and >100,000[35] patients, respectively. Two papers had small patient sizes of ≤100, corresponding to data sizes of 5000–10,000[19,33]. Seven papers did not mention their patient size.

Datasets across studies were mostly public (18 papers) than private (9 papers), with 5 studies using a mixture of the 2 (Table 3[18,21,25,32,35]).

### FL implementation details
Most papers did not specify the FL method used (Fig. 3). Among those that did, horizontal FL was the most frequent (11 papers), with only 2 papers referring to vertical FL[28,30]. The remaining papers did not specify the model-centric technique employed. Horizontal FL was used mainly for classification tasks[17,27,29,36,38], followed by detection[20,23,25], segmentation[25,26] and regression[29].

Most papers did not report their aggregation strategy, with only 5 studies explicitly mentioning the techniques used: federated averaging[15,19,20,32] and consensus model ensemble[14]. Device subtype information was limited, with only 4 papers referencing cross-silo FL[34,35,37,38] and 1 referencing cross-device FL[33]. Only 5 papers explicitly reported their privacy method: differential privacy[29,32], secure aggregation[17], secure multi-party computation[36] and homomorphic encryption[31].

### Evaluating FL scope rigour
We evaluated articles for FL rigour based on 2 objective evaluation criteria: a) whether a comparison framework was used to evaluate FL against central ML baselines on the same datasets, and/or b) whether FL outperformed or showed comparable results to these baselines. The following subsections are organised as follows: we start by describing the 4 most frequent FL scopes identified in our review: "Improving ML model generalisability", "Improving ML prediction", "Data privacy", and "Improving disease understanding". When more than 1 FL scope is involved across papers, all scopes are explicitly detailed at the time each paper is first introduced under the relevant subsection. In other words, when studies address multiple FL scopes, all scopes are explicitly detailed at the first mention of the study within the relevant subsection (e.g., "Improving ML model generalisability"). Conversely, when an FL scope, such as data privacy, is the sole focus of a study, it is discussed in its dedicated subsection. For clarity, in the subsection "Other FL scopes", we separately describe 2 further FL scopes identified in our review: "reduced training time" and "domain adaptation".

### Improving ML model generalisability
FL can enhance model generalisability by using diverse data sources. This subsection highlights articles that demonstrate improved model generalisability through FL, meeting our criteria of comparison to centralised ML baselines and demonstrating superior or comparable performance (see Methods).

Agbley et al. used federated averaging with a pre-trained ResNet on histopathology images to classify breast tumours[19]. To overcome the challenge of dataset variability, the authors integrated various image magnification factors using self-attention. Their FL approach achieved 95.95% accuracy, surpassing various baseline models and demonstrating improved ML generalisability and prediction while protecting data privacy.

In the context of lung cancer, a paper by Zhu et al. presents a novel knowledge-sharing model for pulmonary nodule pre-segmentation and detection from CT data[25]. It uses a 3-stage framework with a) a UNet based mask generator, b) a discriminator with knowledge from electronic medical records and c) a random forest-based lung nodule detector. The system iteratively shares knowledge between a central server and client devices to improve the quality of generated masks and to normalise data distribution, addressing the challenge of non-independent and identically distributed (non-IID) data. Their FL technique outperformed a number of central ML baselines reaching a mean competition performance metric of 89% and a mean Dice score of 76% on non-IID data across each client, therefore improving model generalisability. In[27], the authors explored EHR for lung cancer classification, by developing single and cyclic weight FL using 2 underlying ML models: artificial neural network and logistic regression. By comparing their FL models against the same central ML models across 2 institutions, they showed that FL improved only the artificial neural network-derived results (accuracy ranged from 68-74%). Another collaborative learning method used a federated ResNet model for image classification of lung cancer using large data from 5 institutions with differing labels[35]. In total, they analysed >695,000 thoracic radiographs. The authors proposed a "flexible" FL architecture in which they divide the ResNet model into a classification head and a feature extraction backbone. The feature extraction backbone was shared across all sites, with weights jointly trained under a single FL scheme. Using their "flexible" FL method, the authors showed that model generalisability and classification accuracy were improved against locally trained ResNet models and conventional FL that used only uniformly annotated images.

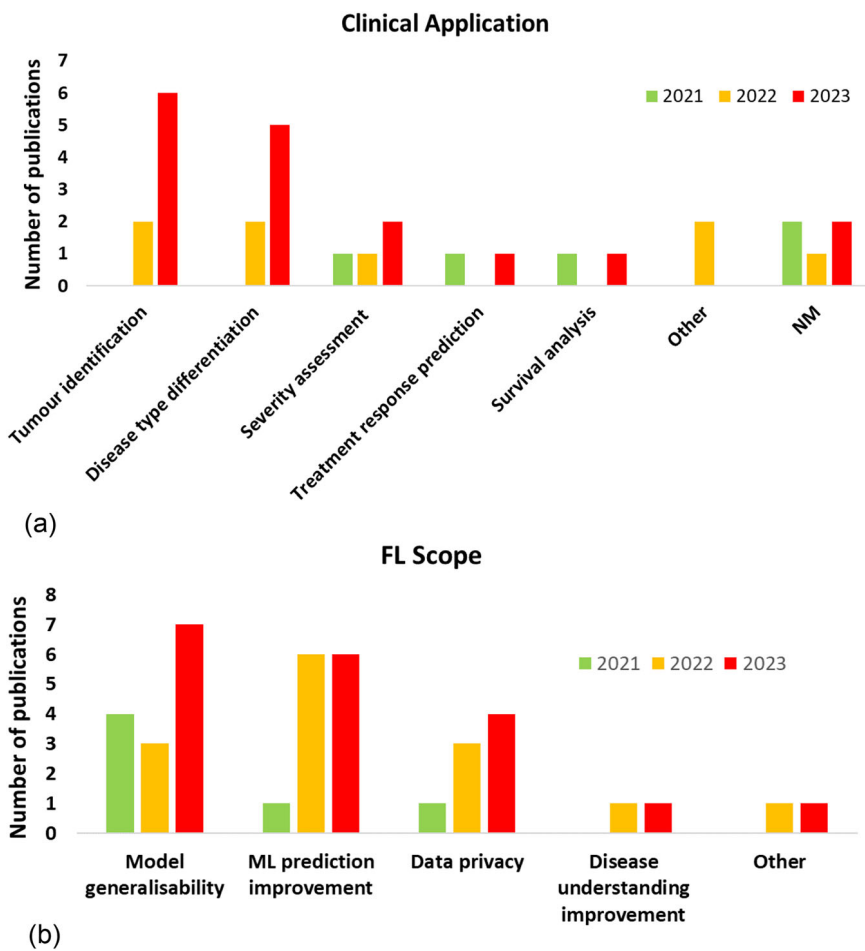

**Fig. 2 | Vertical bar graphs showing research trends for all categories of clinical applications and FL scopes, identified.** Publication record over time for all clinical applications addressed (**a**) and FL scopes (**b**) identified. In **a**, "Other" included side effect prediction (1) and tumour recurrence assessment (1). In **b**, "Other" involved domain adaptation (1) and training time reduction (1). NM: not mentioned; FL: federated learning.

**Table 2 | FL scope and oncology area examined per study**

| FL Scope | Oncology Area | Studies |
| --- | --- | --- |
| Data privacy | Breast | 19,30,33,36 |
|  | Lung | 14,21 |
|  | Prostate | 22 |
|  | Multi-cancer | 23 |
| Domain adaptation | Multi-cancer | 24 |
| ML prediction improvement | Breast | 15,17,19,30,34,36 |
|  | Lung | 20,21,35 |
|  | Prostate | 22 |
|  | Multi-cancer | 18,23,29 |
|  | Multi-disease | 38 |
| Disease understanding improvement | Breast | 15,32 |
| ML model generalisability | Breast | 19 |
|  | Lung | 16,20,25,27,28,35 |
|  | Prostate | 22,26,37 |
|  | Multi-cancer | 23,24 |
|  | Multi-disease | 31,38 |
| Training time reduction | Lung | 21 |

Multi-cancer refers to publications that examine more than 1 oncology area. The term "multi-disease" corresponds to studies in which data from at least 1 of the oncology areas of interest and from other diseases were analysed using FL. FL: federated learning.

In prostate cancer, Yan et al. developed a variation-aware FL (VAFL) method which aimed to assess tumour severity, through MRI classification[22]. The authors introduced VAFL to mitigate cross-client image variation. To perform VAFL, the client with the least complex data was selected to define the target image common space. A privacy-preserving GAN model was trained on these data to synthesise target MR images. A subset of these synthesised MR images was then shared with all clients and each client utilised a modified CycleGAN to map their own images onto this standardised target image space. Using synthetic MRI data, the authors improved both model generalisability and ML classification accuracy in identifying clinically significant prostate cancer, outperforming local classifiers by reaching an accuracy of 98.75%. Data privacy also remained secured. In a multi-cancer work, the authors used a pre-trained ResNet-18 model to perform image detection and a CNN model to classify clinically significant against non-significant prostate cancer (from MRI data) and malignant against benign skin lesions[23]. They report that their FL model secured data privacy, and enhanced both model generalisability and ML classification accuracy, by outperforming locally trained classifiers and other FL model techniques. The diagnostic accuracy ranged from 95.6% to 82.9% on private data and from 88.7% to 73.4% on public data when their FL method was evaluated on 2 up to 32 clients, respectively. Another work focused on developing a federated 3D Anisotropic Hybrid Network for CT image segmentation in prostate cancer[26]. When tested on unseen data, this FL model consistently outperformed 3 local (private) models trained at individual institutions, reaching a Dice score of 89,5% on private data and 88.9% on public data, demonstrating superior generalisability and performance. Gao et al. proposed a novel swarm learning method for MRI and CT image segmentation in multi-disease data (cardiac, brain and prostate

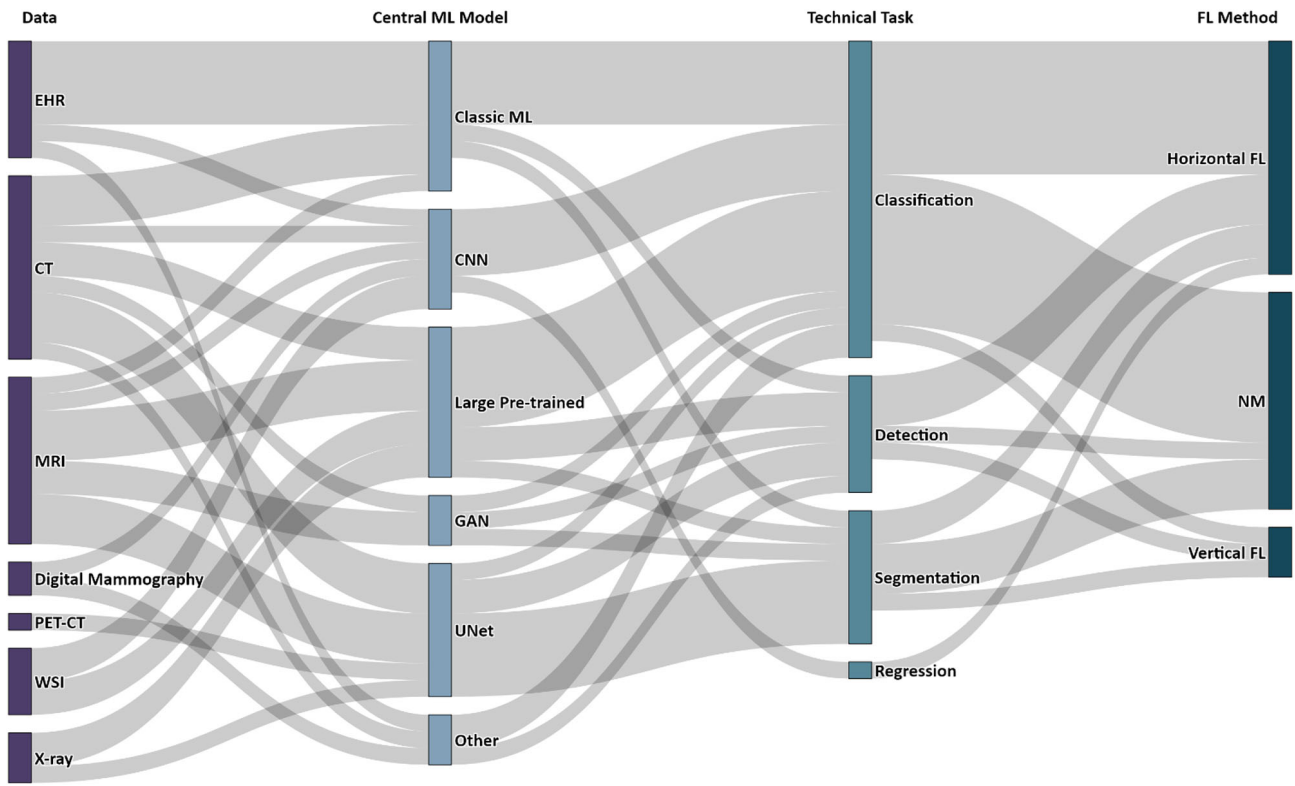

**Fig. 3 | Sankey diagram depicting relationships (combinations) between the following technical aspects extracted across studies (represented as nodes): data, central ML model, technical task addressed and FL method.** The width of each flow is proportional to the quantity being represented: thicker width corresponds to a higher combination prevalence across the reviewed papers, and vice versa. EHR: electronic health records; CT: computed tomography; MRI: magnetic resonance imaging; PET-CT: hybrid positron emission tomography-computed tomography; WSI: whole slide imaging; ML: machine learning; FL: federated learning; NM: not mentioned.

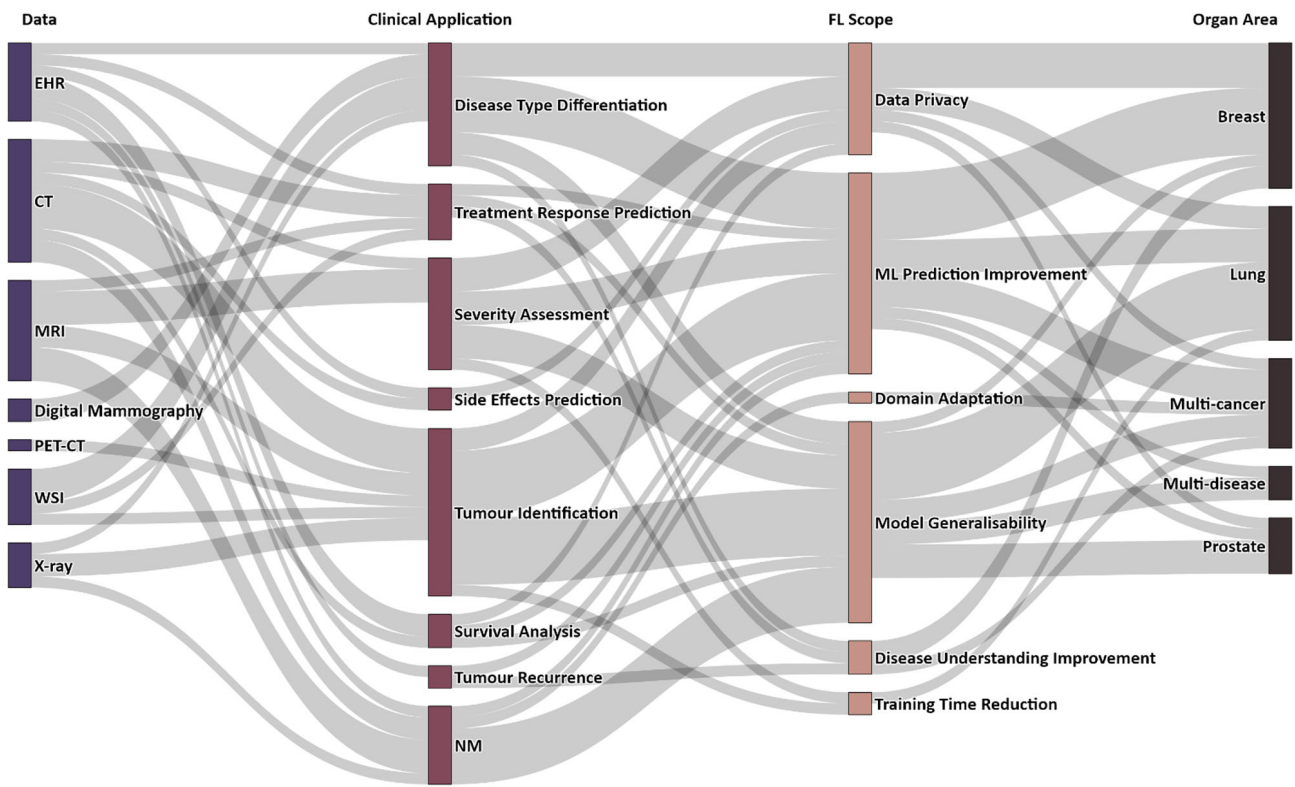

**Fig. 4 | Sankey diagram illustrating combinations between the following application aspects extracted across studies (represented as nodes): data, clinical application, FL scope and organ area.** EHR: electronic health records; CT: computed tomography; MRI: magnetic resonance imaging; PET-CT: hybrid positron emission tomography-computed tomography; WSI: whole slide imaging; FL: federated learning; NM: not mentioned.

Table 3 | Data diversity in terms of patient and data size

| Patient size | Data size | Publications |
|---|---|---|
| 1–100 | 5000–10,000 | 19,33 |
|  | NM | 17 |
| 100–1500 | 1–1000 | 15 |
|  | 1000–5000 | 22 |
|  | 5000–10,000 | 31 |
|  | 10,000–100,000 | 30 |
|  | 100,000–1,000,000 | 20 |
|  | NM | 14,18,26,29 |
| 1500–5000 | NM | 32,37 |
| 5000–10,000 | 10,000-100,000 | 23 |
| >10,000 | NM | 16,27 |
| >100,000 | 100,000–1,000,000 | 35 |
| NM | 1-1000 | 36 |
|  | 1000-5000 | 34 |
|  | 5000-10,000 | 24 |
|  | 10,000-100,000 | 21,25,28,38 |
| Data type | Private | Public |
|  | 14–18,21,25,32,35,37 | 17–25,28,30–36,38 |

Patient size corresponds to the number of individual patients included in the analysis and data size to the number of images or samples analysed. Studies using private and/or public data are also presented at the bottom of the table.

tumours), which could train UNet models using partially labelled images from multiple centres[31]. To tackle inhomogeneous label distributions across centres, they introduced a label skew-awared loss by consolidating global from local knowledge of partial labels. The authors demonstrated Dice scores in the range of 81.1-92.5% on non-IID data (across all disease areas), outperforming other FL methods and showing comparable results to centralised fully-supervised UNet models. In[37], the authors employed a FL approach to train a 3D UNet model with a region-of-interest classification head on diverse annotated prostate MRI data. This collaborative training approach led to substantial improvements in performance compared to local training, boosting lesion classification accuracy by 9.5-14.8% and nearly doubling lesion segmentation accuracy.

**Improving ML prediction**

FL harnesses the collective strength of diverse data sources, mitigating data biases and enabling the model to learn intricate patterns, potentially deriving more accurate predictions. In the following paragraphs, we detail articles which were designed to improve ML predictions through FL. Since there were FL studies that aimed to improve both ML model predictions and generalisability, we have already detailed these overlaps in the previous subsection[19,21,23,35]. Here below, we highlight the remaining articles that satisfied our 2 criteria in terms of improving ML prediction: comparison to centralised ML and whether FL outperformed the centralised ML baseline(s) (see Methods).

Breast cancer was the most frequent area in which FL was used for ML performance improvements[15,17,19,30,34,36]. Study[15] applied federated averaging and multiple instance learning to WSI and clinical data, to predict the histological response to chemotherapy in early-stage triple-negative breast cancer. The authors aimed to improve ML predictions as well as disease understanding and achieved a mean AUC of 66% (range: 57–78%), outperforming central ML baselines. Study[17] focused on federating extremely randomised trees, to analyse distributed structured health data for disease classification, reporting an accuracy of 95.3% and an F1 score of 95.4%. It showed comparable results to central ML baselines. Another work expanded the application of privacy-preserving FL to survival analysis, assessing its potential in breast cancer genomics, with reported accuracies ranging from 81-92.5% across all 4 datasets explored[30]. The proposed FL outperformed all central ML baselines. In[34], the authors used a DenseNet feature extractor which fed an RNN-based classifier for breast cancer classification, from digital mammography data. The authors trained the model using a hybrid optimisation algorithm, achieving an accuracy of 94.22%, outperforming 3 central ML and 1 FL baselines. Another breast cancer work focused on enhancing model prediction and protecting data privacy by using a simple neural network classifier on clinical datasets, reaching an accuracy of 99% and outperforming a number of central ML baselines[36].

In lung cancer, we found 2 papers that applied FL to enhance ML predictions[20,21]. The first study showcased an FL model employing a 3D ResNet18 for lung nodule detection, achieving an accuracy of 83.41%[20]. Although the proposed FL model was compared against previous FL methods, there was no comparison against central ML baselines. Study[21] combined FL with blockchain to create a collaborative, privacy-preserving model for lung cancer classification, demonstrating a high detection accuracy of 99.69% which outperformed central ML methods. The authors also reported that their FL method led to reduced training time, compared to central ML models.

In a multi-cancer study (including breast cancer)[29], the authors performed differential privacy to predict tumour recurrence from clinical data, by using a CNN model. The authors focused to improve ML predictions and disease understanding, demonstrating high FL accuracy ( >90%) and outperforming central ML baselines.

**Data privacy**

Although data privacy is a core aspect of FL, only 8 papers explicitly mentioned it within their scope. Most of these papers were covered in the subsections "Model generalisability"[19,22,23] and "Improving ML prediction"[21,30,36], as their data privacy considerations were intertwined with those topics. In 2 publications, data privacy was addressed as a standalone focus and these are further discussed below[14,33].

Peta and Koppu developed an automated breast cancer classification system using FL combined with deep learning to enhance diagnostic accuracy from histopathological images[33]. To ensure data privacy, they employed an Extended ElGamal Image Encryption algorithm to encrypt the medical images before sharing them within the FL framework. The encrypted images were stored and processed within a FL flower framework, which further enhanced data security during transmission. The authors compared the performance of their proposed Convolutional Capsule Twin Attention Tuna Optimal Network against existing deep learning models (i.e., CNN, BiLSTM, DNN, CapsuleNet), using the public BreakHis dataset. They reported superior performance with an accuracy of 95.68%, outperforming all previous centralised ML models.

In their lung cancer classification work, Field et al. developed a federated consensus model ensemble using logistic regression on private clinical data[14]. The authors prioritised data privacy through a FL infrastructure, enabling collaborative model development without direct data sharing. While specific security methods were not detailed, this decentralised approach inherently minimises privacy risks by keeping patient data within each hospital and limiting information exchange to model parameters or aggregated statistics. Their FL method demonstrated comparable accuracy to their central baselines, reaching a mean AUC of 70%.

**Improving disease understanding**

Three of the papers reviewed used FL to improve disease understanding: 2 papers were about breast cancer[15,32] and 1 was on multiple cancers[29]. Two of these papers have already been detailed in the "Improving ML prediction" subsection[15,29]. In the remaining paper, the authors used federated averaging and a memory-aware CNN to refine breast cancer classification from digital mammography data[32]. This approach was shown to prioritise challenging instances that often experience prediction inconsistencies, thereby contributing to a more refined understanding of disease characteristics and diagnostic accuracy.

**Other FL scopes**

Two further FL scopes, reduced training time[21] and domain adaptation[24], were each the subject of 1 paper in our review. The study by Heidary et al. has been previously described in the "Model generalisability" and "Improving ML prediction" subsections[14]. Although the authors in[24] describe that their FL methods outperformed conventional deep learning models, there was no such comparison identified in their study. Therefore, their work did not meet our criteria for rigorous FL research and are not further described herein.

## Discussion

In this systematic review, we examined FL methods in breast, lung and prostate cancer from 2020–2023. Distinct from previous surveys that focused on the theoretical and technological aspects of FL[1,6,7,9,12,39], our analysis investigates its practical implementation and impact in real-world breast, lung and prostate cancer settings. We examined the scope and evaluated the methodological rigour and effectiveness of FL in improving major ML domains such as model generalisability, predictive accuracy and data privacy, by comparing FL to centralised models and assessing its scientific and clinical impact. Most papers (18 out of 25 reviewed) met our objective criteria for FL rigour, including extensive comparisons against centralised ML baselines. Notably, FL methods outperformed centralised ML in 15 papers and showed comparable results in 3 papers. A diverse range of FL/ML techniques and their key clinical applications were also comprehensively explored and revealed. This review contributes meaningful insights into the real-world application of FL in cancer research, supporting the transition from promising proof-of-concepts to widespread implementation in clinical settings.

In light of recent advances in FL applications within healthcare, our systematic review bridges significant gaps identified in the current literature. Several recent studies provide valuable insights into the implementation of FL for specific diseases or data types and demonstrate FL's potential to enhance data privacy and model accuracy[10,11,40–46]. However, these studies often focus on singular aspects of FL implementation or are limited to specific healthcare applications. Our review extends this analysis by systematically evaluating a broader range of FL applications across multiple types of cancer—breast, lung, and prostate —thereby offering a more comprehensive overview of FL's capabilities and challenges in oncology. For instance, unlike previous review papers which primarily focus on data privacy[40,41] or general FL applications in healthcare[10,11,42–45], none of which specifically address oncology, our review specifically examines the real-world impact of FL on model performance and generalisability within the context of oncology, with a rigorous evaluation of its effectiveness in enhancing model performance and generalisability across diverse oncological datasets. In integrating these diverse perspectives, our review not only underscores the multi-dimensional benefits of FL in healthcare but also highlights critical areas for future research and development, ensuring that FL can meet the complex demands of modern oncology.

Among studies meeting our rigorous FL criteria, a diverse range of ML models were employed, including Large Pre-trained models, Classic ML, UNets and CNNs. Classification was the most common task, followed by segmentation and detection. While tumour identification and disease type differentiation were the dominant clinical applications, all clinical applications were represented in this group of studies. Improving ML generalisability and predictive accuracy were the primary focus of most studies meeting our FL criteria. Nevertheless, all FL scopes were represented, except for domain adaptation. Breast and lung cancer were the primary areas of focus, but studies on all organ areas were observed. Of these studies, 2 involved large patient populations (>1500 patients)[23,32] and 2 used very large datasets with over 20,000 and 900,000 patients, respectively (Supplementary Table 1)[27,35]. The majority (6 papers) used moderate to large patient sizes (300-1,500 patients)[14,15,18,22,26,29–31]. These FL techniques can potentially have merit to be further validated as generalised solutions for real-world cancer research.

By training models on large, diverse datasets from multiple sources, FL can uncover hidden patterns that may not be evident in smaller, isolated datasets. This can lead to the development of more precise diagnostic and treatment strategies tailored to specific patient groups (precision medicine)[47,48]. FL can further support precision medicine, by potentially integrating multi-modal information from genetic, clinical, and (bio-) medical imaging data across institutions[49–56]. In fact, among studies that satisfied our FL rigour criteria, there were 7 papers which performed multi-modal ML analysis, combining information from various sources such as EHR, (bio-)medical imaging and/or genomics[14,15,24,25,29–31]. Of these studies, 1 focused on predicting response to treatment using WSI and clinical data[15], while the others aimed to develop diagnostic or identification methods. Thus, multi-modal FL offers a unique advantage for advancing precision medicine by integrating diverse patient-level data types across institutions, potentially enhancing patient outcomes.

Our analysis reveals a strong emphasis on generalisable models within FL frameworks across multiple oncology domains. Most of these publications met our FL rigour criteria[19,22,23,32,35,37]. The inclusion of heterogeneous data across institutions can potentially enhance the ability of ML models to learn more representative patterns, reduce overfitting and perform robustly across various clinical settings[47]. We also observed a large variability in data types across studies in this area, spanning from histopathological images to MRI, CT and EHR data. This shows that FL has considerable scope for developing generalisable ML models with broad applicability. Of note, 2 of these publications report results from non-IID data, which is central to ML model generalisability[2,47]. Next to generalisability, learning from broad and heterogeneous data patterns across institutions can potentially improve ML performance. We observed a substantial focus on improving ML performance across oncology domains and data types. Most of these papers satisfied our FL rigour criteria[15,29,30,34,35]. The surge in publications from 2021 to 2023 underscores a growing awareness of these FL opportunities within the cancer research community.

Our findings highlighted potential advancements in evaluating tumour severity[37] and enhancing disease understanding and characterisation[15,29,32]. These advancements can make clinical workflows more efficient by potentially assisting in patient stratification, risk assessment as well as enhancing early diagnosis and treatment strategies. FL offers a notable advantage in accelerating ML model training through parallel processing, data privacy (eliminating the need for data sharing) and fast convergence due to model averaging[1,6,12]. A lung cancer study reported shortened training times without compromising data integrity[21], but other studies did not evaluate this aspect. These benefits may be offset by communication overheads due to sharing model updates (especially with large models or frequent updates) and the complexities of coordinating heterogeneous devices and privacy preservation techniques[39]. Further work is required to potentially demonstrate the benefits of FL in expediting ML model training and data-driven decision making.

Moreover, there are limitations and challenges that must be considered. The major limitation identified was the lack of FL implementation details across most publications. Specifically, only 13, 5 and 5 out of the 25 papers explicitly described the FL method, aggregation strategy and device subtype used, respectively (see Results). Of note, only 5 papers detailed their privacy method. Although most FL papers involved at least 1 public dataset, 4 papers involved only private data (Table 3). Moreover, among the 18 papers that met our FL rigour criteria, only 8 provided accessible code[14,15,17,18,31,32,35,37]. These limitations hamper reproducibility and wider FL model adoption. We strongly advocate for increased transparency and open access practices to foster wider exploration and validation of these approaches in future large-scale cancer research. Comprehensive documentation of FL methods and code is important to ensure that findings can be reliably reproduced and externally validated.

Another major limitation was the use of datasets with low or moderate patient diversity in some studies (Table 3, Supplementary Table 1), which may be associated with limited generalisability, reduced robustness and susceptibility to biases. In addition, 7 papers did not meet our FL rigour criteria, which means that FL was not cross-validated against central ML baselines. Hence, the presence of systematic biases and the effectiveness of these FL methods in achieving their intended scope and clinical application remain unclear.

The underlying ML models used varied considerably even across the same technical tasks (Fig. 3, Table 1). For instance, all ML model families were used to address either classification or detection tasks. This heterogeneity in both problems (data, oncology domains) and scientific solutions (ML models, FL methods) makes establishing best practices and method scalability challenging. In addition, many studies did not use a consistent set of metrics (Table 1). For example, while some studies report improvements over central ML models using accuracy and AUC metrics, others relied on precision, recall and F1-scores without clear justification. Another potential hurdle is the lack of universally accepted benchmark datasets for evaluating FL. Benchmark data with appropriate ground truths and evaluation metrics[57], are essential for unbiased comparison of different FL approaches. The lack of such datasets complicates transparent FL evaluations. However, most FL papers in our review involved at least 1 public dataset in their evaluation, which can help to establish frameworks for thorough benchmark studies.

Data privacy is a crucial concern in FL, and our review highlighted various techniques such as differential privacy and secure computation being commonly used to safeguard patient information[1,6]. These techniques ensure that privacy is maintained across FL systems. In our review, most of the papers reporting implementation details, focused on federated averaging[15,19,20,32] which mitigates overfitting and enhances model generalisation by aggregating models trained on diverse local datasets[58].

While foundation models, pretrained on vast amounts of diverse data, offer a powerful starting point for various machine learning tasks[59], FL provides unique advantages in healthcare domains like oncology, where data sensitivity and privacy are paramount. FL enables collaborative model training without direct data exchange, ensuring compliance with strict privacy regulations and fostering trust among institutions. Furthermore, as demonstrated in our review, FL can effectively address the inherent heterogeneity of oncological data across institutions, leading to models with improved generalisability and enhanced performance, as evidenced by the studies that met our criteria of comparison against centralised ML baselines. Future research should explore the optimal integration of foundation models within FL systems, carefully balancing their benefits with the need to preserve privacy and address data heterogeneity in real-world oncological settings.

The future development and effective application of FL in oncology will rely on sophisticated aggregation techniques that can handle the heterogeneity and variability inherent in clinical datasets. Based on our domain expertise, we propose that future research prioritise advanced aggregation methods such as FedProx, FedNova, FedDyn, and Fed-ROD to overcome challenges related to data heterogeneity, client drift, and fairness, which are particularly relevant in oncology due to the variability of patient populations and data sources. FedProx may be promising for heterogeneous oncology data, since it regularises the local objective function and expedites training across diverse data distributions[60]. FedNova further refines this optimisation by normalising contributions based on local training steps[61], whilst FedDyn adapts the learning process to the unique characteristics of each local dataset[62]. Unlike other FL methods that prioritise either global or local model performance, Fed-ROD uniquely addresses fairness by optimising for both simultaneously, ensuring equitable outcomes for all clients regardless of the data distribution[63]. Moreover, the FedOpt framework offers flexibility in integrating various FL optimisation algorithms, allowing adaptation to the specific constraints and requirements of medical research[64]. Federated transfer learning, where pre-trained models are fine-tuned with federated data, can also potentially enhance ML predictions and disease understanding by improving adaptation to new oncology data[65]. Future work could investigate further advanced FL methods such as the aforementioned to harness data heterogeneity for improved model generalisation, performance and fairness.

In conclusion, this comprehensive review serves as a foundational resource for the broader oncology community, demonstrating the burgeoning potential of cutting-edge FL to revolutionise breast, lung and prostate cancer research by leveraging diverse, real-world datasets while maintaining patient privacy. It also aligns perfectly with the overarching goals of the OPTIMA IMI2 project and provides crucial insights that will guide future research. Despite current challenges in reproducibility, standardisation and methodology, the clear advantages of FL in enhancing model generalisability, performance and addressing clinical needs across various cancer types, highlight its immense promise for harnessing diverse real-world clinical data and transforming cancer care.

## Methods
### Literature review strategy
A comprehensive literature search was conducted following the Preferred Reporting Items for Systematic Reviews and Meta-Analyses (PRISMA) guidelines (Fig. 5)[13]. We systematically reviewed the literature published on FL techniques in breast, lung, and prostate cancer ML analysis, from January 1, 2020 to September 1, 2023. This period captures recent developments in FL, reflecting its transition to an established practice amid tightening data privacy regulations[6,7,9,12]. Despite initiating our literature review in early 2020, we found no publications matching our criteria until 2021 (Figs. 1, 2). Database searches included PubMed, Scopus, and Web of Science, utilising a combination of keywords and medical subject headings (MeSH) terms related to oncology, FL and ML. An expert librarian from the University of Oxford specified the keyword space.

Initial searches using broad keywords in abstracts, titles, and manuscript keywords related to oncology and ML yielded 5,766 papers. These keywords were: Oncology OR Cancer OR Carcinoma OR Malignant OR Neoplasm OR Tumor OR Tumour; Machine Learning OR Deep Learning OR Convolutional Neural Network OR CNN OR Generative Adversarial Network OR GAN OR Variational Autoencoder OR VAE OR Diffusion OR Transformer.

Subsequently, to focus the search, we mined only publications relevant to the 3 oncology areas of interest, FL and all possible patient-level data types involved in oncology. Specifically, we added the following keywords in the abstract, title, and manuscript keywords: Breast OR Mammary OR Prostate OR Lung Oncology OR Cancer OR Carcinoma OR Malignant OR Neoplasm OR Tumor OR Tumour; Federated Learning; Real World Evidence OR Real World Data OR Medical Imaging OR Magnetic Resonance Imaging OR Computed Tomography OR Positron Emission Tomography OR Ultrasound OR Echo OR Digital Breast Tomography OR Digital Pathology OR Genetic OR Genomics OR Transcriptomics OR Electronic Health Records OR Clinical Data OR Hospital Data OR Primary Care Data OR Secondary Care Data OR Computational Biology. By adding these terms, we removed all irrelevant publications to the 3 oncology areas, FL and patient-level data types, leaving 81 papers for screening (Fig. 5).

Non-journal publications and duplicates were removed by 6 authors (AA, MA, EG, SB, SAH, GP), resulting in 40 papers. These authors screened further for relevance focusing on FL methods in the specified cancer types using ML for analysis. We excluded non-English publications, studies outside the specified data types and oncology areas, and studies not focusing on FL applications, leaving 31 papers for full-text review. Following full-text review by the same 6 authors (in groups of 2 researchers per publication) and inclusion/exclusion criteria, we removed 6 more papers. Conflicts between authors were resolved through consensus with the rest of the review team. In total, 25 journal papers were included in our review analysis.

### Review aspects
During full paper review, we considered the following aspects: (1) year of publication; (2) central ML technical task addressed; (3) clinical application addressed; (4) central ML model architecture used; (5) FL method; (6) aggregation strategies for FL; (7) device types; (8) datasets used for analysis (imaging, electronic health records, and other); (9) privacy method; (10) FL evaluation; (11) scope of FL; (12) oncology area. We also evaluated data diversity by (13) patient size and (14) data size. Furthermore, to assess whether it is possible to reproduce these FL techniques, we considered (15) data type per publication (private, public or both).

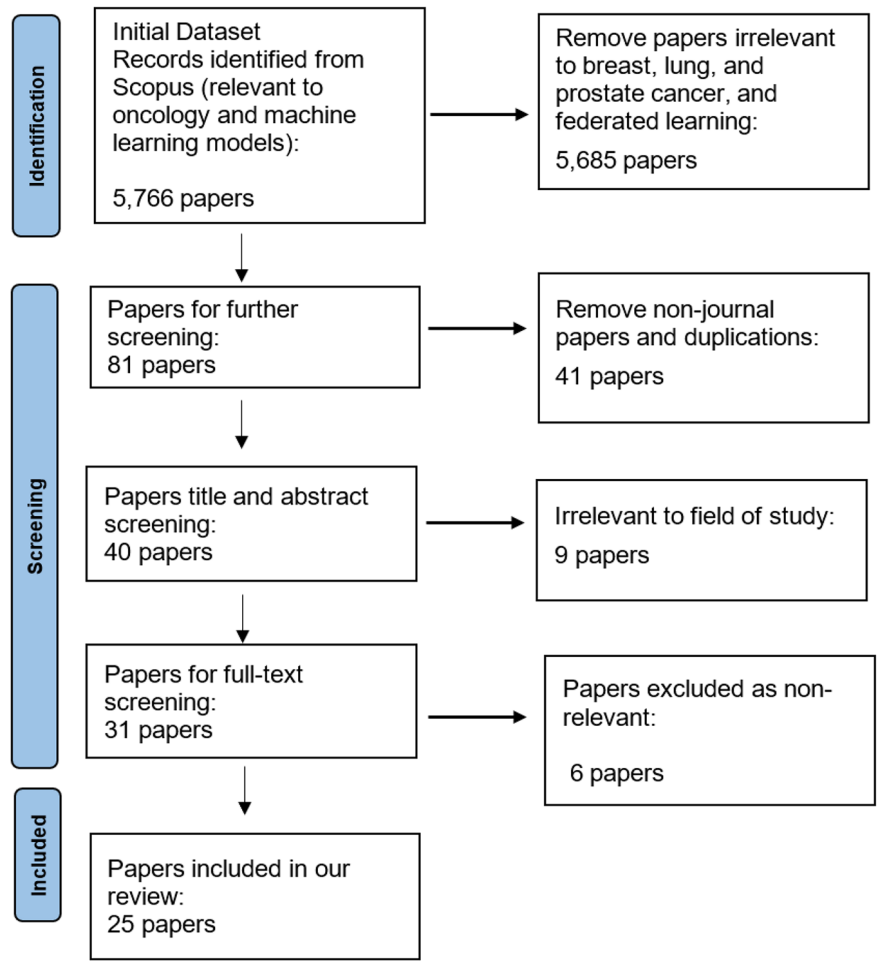

**Fig. 5 | PRISMA flow of the systematic review process.** The flow presents inclusion and exclusion of papers at each review stage.

For clinical applications (3), we categorised: disease type differentiation, tumour identification, treatment response prediction, severity assessment, side effect prediction, survival analysis, and assessment of tumour recurrence. To improve clarity, we categorise all published work on cancer diagnosis against benign tumours and diseases other than cancer as "disease type differentiation", on tumour classification, segmentation or detection from (bio-)medical imaging data as "tumour identification" and on staging as "severity assessment". For ML models (4), we categorised: Classic ML: logistic regression, support vector machine, extremely randomised tree, random survival forest, random forest; Convolutional Neural Network (other than UNets and pre-trained models) (CNN ALL): CNN (Encoder-Decoder, E-D), 2D CNN (E-D), 3D CNN (E-D), CNN (E-D) with multiple instance learning; Large Pre-trained (ALL): 2D ResNet, 3D ResNet, ResNet-201, ResNet-18, Mobilenet-v3, Xception-v3; GAN (ALL): GAN, WGAN, CycleGAN; UNet (ALL): 2D UNet, nnUNet, 3D UNet; Other: capsule neural network, recurrent neural network and region-based CNN.

In (5), we considered the 2 major FL methods: model-centric and data-centric[1]. In (6), aggregation strategy refers to the method by which updates or model parameters from participating devices are combined to generate a global FL model[1,6,7,9,12]. This process involves aggregating local model updates or parameters, to balance the contributions of individual devices while preserving the privacy of the local data. Aggregation strategies play a pivotal role in FL by ensuring that the global model accurately represents the collective knowledge of all participating devices[1,6]. For (7), the term "device types" refers to different categorisations of "central" devices based on their data distribution and functionality within the FL framework[1,6]. In our reviewed work, we identified 2 device types: cross-device and cross-silo. In (11), the "scope of FL" encompasses all major end goals of scientific and clinical impact identified in our review: ML model generalisability (to unseen data instances from various local devices), ML model prediction improvement (by potentially learning from a broad range of data patterns and insights across devices), data privacy, disease understanding improvement (through the analysis of multiple device data), domain adaptation and training time reduction. Note that apart from data privacy, all other scopes benefit from increased data heterogeneity and patterns provided by FL/ML.

We assessed FL rigour based on 2 objective criteria across all FL scopes: a) whether a comparison framework was involved to evaluate the proposed FL technique against central ML baselines developed on the same datasets (either directly evaluated in the study or by reporting literature values); and b) if they outperformed or showed comparable results to these baselines.

## Data availability
The data extracted in the study is available from the corresponding authors upon reasonable request.

## Code availability
No code was used for this study.

## Acknowledgements
We are grateful to Shona Kirtley, expert librarian from the University of Oxford, for her invaluable assistance in defining the keyword space. We thank the OPTIMA consortium for the support. OPTIMA is funded through the IMI2 Joint Undertaking and is listed under grant agreement No. 101034347. IMI2 receives support from the European Union's Horizon 2020 research and innovation programme and the European Federation of Pharmaceutical Industries and Associations (EFPIA). IMI supports collaborative research projects and builds networks of industrial and academic experts in order to boost pharmaceutical innovation in Europe. The views communicated within are those of OPTIMA. Neither the IMI nor the European Union, EFPIA, or any Associated Partners are responsible for any use that may be made of the information contained herein.

## Author contributions
A.A., G.P., S.B., M.A., E.G., and S.K. were involved in conceptualization, study design and methodology. A.A, G.P., S.B., M.A., E.G., S.A.H, and S.K., performed screening of data sources, data extraction, and data analysis. A.A. and G.P. wrote the original draft of the manuscript. A.A., G.P., S.B., M.A., G.Y., C.B., N.D., G.A.K., C.C., S.K. revised and contributed to the final draft. M.B., H.K., and P.L. contributed to the interpretation of the findings. G.P. supervised the project. All authors have read and approved the manuscript.

## Competing interests
G.P., S.B. and H.K. are full-time employees of Pfizer and hold stock/stock options. C.B. is a full-time employee of Flower. P.L. has/had minority shares in the companies Radiomics SA, Convert Pharmaceuticals SA, Comunicare SA, LivingMed Biotech srl and Bactam srl. The other authors do not have any financial or non-financial competing interests to declare.

## Additional information
**Supplementary information** The online version contains supplementary material available at
https://doi.org/10.1038/s41746-025-01591-5.

**Correspondence** and requests for materials should be addressed to Anshu Ankolekar or Giorgos Papanastasiou.

**Reprints and permissions information** is available at
http://www.nature.com/reprints

**Publisher's note** Springer Nature remains neutral with regard to jurisdictional claims in published maps and institutional affiliations.

# OPTIMA Consortium

Anshu Ankolekar[1] ✉, Sebastian Boie[2], Maryam Abdollahyan[3], Emanuela Gadaleta[3], Seyed Alireza Hasheminasab [4], Michael Bussmann[8], Claude Chelala[3], Sara Khalid[4], Hagen Kruger[2], Philippe Lambin[1] & Giorgos Papanastasiou[9] ✉